\documentclass[conference]{IEEEtran}

\IEEEoverridecommandlockouts
\usepackage{cite}
\usepackage{amsmath,amssymb,amsfonts}
\usepackage{algorithmic}
\usepackage{graphicx}
\usepackage{textcomp}
\usepackage{xcolor}

\usepackage{caption}
\usepackage{subcaption}

\usepackage{algorithm}

\usepackage{array}

\usepackage[a4paper, total={6in, 8in}]{geometry}

\def\BibTeX{{\rm B\kern-.05em{\sc i\kern-.025em b}\kern-.08em
    T\kern-.1667em\lower.7ex\hbox{E}\kern-.125emX}}

\begin{document}

\title{Anomaly Detection of Underwater Gliders\\
Verified by Deployment Data
\thanks{
The research work is supported by ONR grants  N00014-19-1-2556 and N00014-19-1-2266;  AFOSR grant FA9550-19-1-0283; NSF grants GCR-1934836,  CNS-2016582 and ITE-2137798; and NOAA grant NA16NOS0120028.
}
}

\author{
\IEEEauthorblockN{Ruochu Yang}
\IEEEauthorblockA{\textit{School of Electrical and Computer Engineering} \\
\textit{Georgia Institute of Technology}\\
Atlanta, United States}
\and
\IEEEauthorblockN{Mengxue Hou}
\IEEEauthorblockA{\textit{College of Engineering} \\
\textit{Purdue university}\\
West Lafayette, United States}
\and
\IEEEauthorblockN{Chad Lembke}
\IEEEauthorblockA{\textit{College of Marine Science} \\
\textit{University of South Florida}\\
St.Petersburg, United States}
\and
\IEEEauthorblockN{Catherine Edwards}
\IEEEauthorblockA{\textit{Skidaway Institute of Oceanography} \\
\textit{University of Georgia}\\
Savannah, United States}
\and
\IEEEauthorblockN{Fumin Zhang}
\IEEEauthorblockA{\textit{School of Electrical and Computer Engineering} \\
\textit{Georgia Institute of Technology}\\
Atlanta, United States}
}

\maketitle

\begin{abstract}
    This paper utilizes an anomaly detection algorithm to check if underwater gliders are operating normally in the unknown ocean environment. Glider pilots can be warned of the detected glider anomaly in real time, thus taking over the glider appropriately and avoiding further damage to the glider. The adopted algorithm is validated by two valuable sets of data in real glider deployments, the University of South Florida (USF) glider  Stella and the Skidaway Institute of Oceanography (SkIO) glider Angus.
\end{abstract}


\section{Introduction}
\label{intro}
Autonomous underwater profiling gliders, hereafter referred to as gliders, have been widely used in oceanography for a range of applications, including harmful algal bloom monitoring, improving hurricane prediction, passive and active acoustics, ocean observing systems, and targeted science questions \cite{uncrewedocean, 10.3389/fmars.2019.00422, weisberg2019coastal, cauchy2018wind, whaledetection}. Similar progress has been made to improve the performance and value of the data collected through coordinated control of vehicles over long durations \cite{doi:10.1080/00207170701222947, 4476150, hou2020}. Despite these advances, unpredictable events like shark or vessel strikes \cite{inproceedings}, wing loss, or interaction with echneids (remoras) can lead to abnormal flight behavior, prematurely end a glider's mission, or even result in the loss of the vehicle. Human pilots cannot always directly track gliders' abnormal behavior from the subsets of measured data that are telemetered to shore, especially when gliders confront external disturbances \cite{gertler2017fault, chen2012robust}, and sending more data or adding monitoring devices to detect abnormal behavior is not always feasible or helpful. Therefore, anomaly detection algorithms can be developed to aid human pilots, and their design must rely on a limited subset of glider data.

Anomaly detection has been studied extensively in the robotics community \cite{raanan2016automatic, park2016multimodal, isermann2005model}. Specifically in the domain of marine robots, some anomaly detection algorithms focus on individual components that frequently degrade like propellers, thrusters, and rotors \cite{fagogenis2016online, sun2016thruster, caiti2015enhancing, caccia2001experiences}. Other anomaly detection algorithms utilize robot motion like monitoring pitch angle and depth angle to detect possible deviation from expected motion \cite{7404462}.  However, research to date has not yet resolved issues such as inaccessible hardware dynamics, incapability of performing online detection, and lack of experimental verification. Detected anomalies must also be checked against the case of false alarm \cite{chandola2009anomaly}. One common scenario occurs when flow speed exceeds the maximum robot speed; the performance of marine robots is degraded in terms of speed over the ground \cite{doi:10.1080/00207170701222947}, but the anomaly occurs due to environmental conditions rather than  flight performance of the robot itself. Therefore, it is significant to compare marine robot speed with ocean flow speed;   this comparison is not trivial since flow speed is not directly measured by the glider but is instead estimated through dead reckoning. 

Controlled  Lagrangian particle tracking (CLPT) is a theoretical  framework  that can  simultaneously estimate  ocean flow speed and robot speed  \cite{szwaykowska2017controlled}. Based on these two estimates, it is feasible to detect abnormal motion while providing robust false alarm avoidance by excluding unexpected ocean flow. The anomaly detection algorithm applied in this paper is adopted from the work \cite{cho2021learning}, which uses trajectory data and heading angle data that are available and generic in all types of gliders. This similar technique of leveraging glider navigational data can also be seen in previous work \cite{morris2008survey, lei2016framework, rosen2012line}. Given a trajectory and  heading angles, the adopted algorithm generates both the estimated flow speed and the estimated glider speed in real time. The estimated glider speed can be compared with the normal speed range to check for anomalies, and the  estimated flow speed is compared with the glider-estimated flow speed to check if the detected anomaly is a false alarm.

In general, anomaly detection is difficult to validate in field experiments. In abnormal conditions, gliders are likely to encounter issues of GPS search, thus imposing risk and cost to human rescue. Anomalies occur unexpectedly, there is lack of ``ground truth" to determine the major factor causing the anomaly.  Further, anomalies can accumulate over time, and it can be difficult for glider pilots to notice in real time even with close monitoring during field experiments. Given only a subset of data, small changes in flight performance can be too subtle for pilots to identify. The main contribution of this paper is to verify the anomaly detection algorithm in \cite{cho2021learning} using real-life glider deployment data for the first time. The algorithm can infer whether or not the glider may have had a shark hit, wing loss, or remora attachment simply from a subset of flight data rather than large amounts of data.  The University of South Florida (USF) glider Stella and the Skidaway Institute of Oceanography (SkIO) glider Angus provide two valuable sets of experimental data, in which anomalies can be attributed to marine bio-hazards and strong evidence is available to confirm their cause.  Both data sets show promising detection results to validate the anomaly detection algorithm in \cite{cho2021learning}.

 This paper is organized as follows. Section \ref{experiment setup} describes the working principles of Slocum glider and the experiment setup of two coastal glider deployments in the Gulf of Mexico and south east Atlantic Ocean. Section \ref{anomaly detection algo} illustrates the framework of the anomaly detection algorithm. Section \ref{experimental results} verifies the algorithm by showing results of detected anomalies. Section \ref{conclusion} provides conclusions and future work.

\section{Glider and Experiment Setup}
\label{experiment setup}

This section introduces the Slocum glider, its method of navigation, and the details of deployments of two gliders, Stella and Angus, that were subjected to anomalous conditions while deployed. 

\subsection{Slocum Glider}
The Slocum glider is an autonomous underwater vehicle (AUV) that moves by changing its buoyancy and center of gravity \cite{robustandready}. Capable of collecting data in the ocean for long duration missions of up to 6 months without a recharge, gliders typically travel at low speed (approximately $0.25–0.35 m \cdot s^{-1}$).  During the deployment, the glider surfaces at pre-defined intervals to transmit flight and science data, as well as receive updates and other instructions via satellite connection.  In order to minimize time at the surface, where the glider is vulnerable to ship traffic and other dangers, only a small portion of heavily subsetted flight data are telemetered to shore in small binary data (SBD) files.  For example, in The trajectory and heading information of sbd file data is processed in real time by the anomaly detection algorithm presented in Section \ref{anomaly detection algo}. After the mission is over, all the full data sets are downloaded off the glider into Dinkum binary files (DBD and EBD for flight and science data, respectively).  The anomaly detection algorithm is validated in Section \ref{experimental results} by comparing the anomaly detected from the real-time SBD file date with the anomaly directly shown from the full DBD file data.

While in mission, the glider estimates its horizontal position using pitch and depth data through dead reckoning. From one surfacing to the next, the deviation the glider makes from the dead reckoned estimate is attributed to the influence of flow.  The difference  between these two positions are used to estimate the average flow speed along the glider trajectory.  GPS positions at the beginning and end of the surfacing are used to adjust the effect of surface wind drift on estimated flow velocity. This glider-estimated flow is checked against the flow estimate given by the anomaly detection algorithm in Section \ref{anomaly detection algo}.

\subsection{Field Deployments}

\begin{figure}[htbp]
       \centerline {\includegraphics[width=0.4\textwidth]{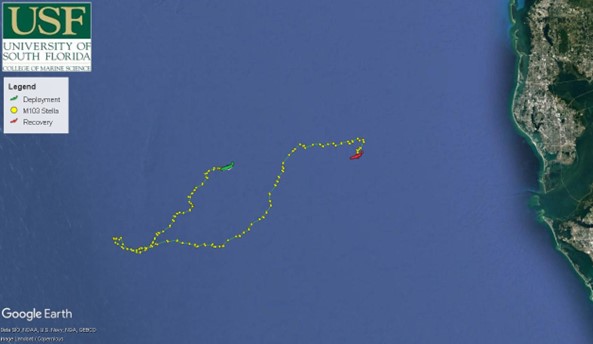}}
        \caption{Google Earth trajectory of the 2021 Stella deployment.}
        \label{stella google earth traj}
    \end{figure}

Data from two gliders will be used for field validation of the detection anomaly algorithm: the glider Stella, operated by the University of South Florida, and the glider Angus, operated by Skidaway Institute of Oceanography, part of the University of Georgia.  
Stella ($\#$772) was deployed by the University of South Florida in the Gulf of Mexico on the West Florida Shelf near $(27.6970^{\circ} N, 83.8778^{\circ} W)$ on July 01, 2021 UTC, transited offshore and then back inshore, and ended at the location $(27.7318^{\circ} N, 83.4136^{\circ} W)$ on July 19, 2021 UTC. The Google Earth trajectory of Stella is shown in Fig.~\ref{stella google earth traj}.

\begin{figure}[htbp]
       \centerline {\includegraphics[width=0.4\textwidth, height=4cm]{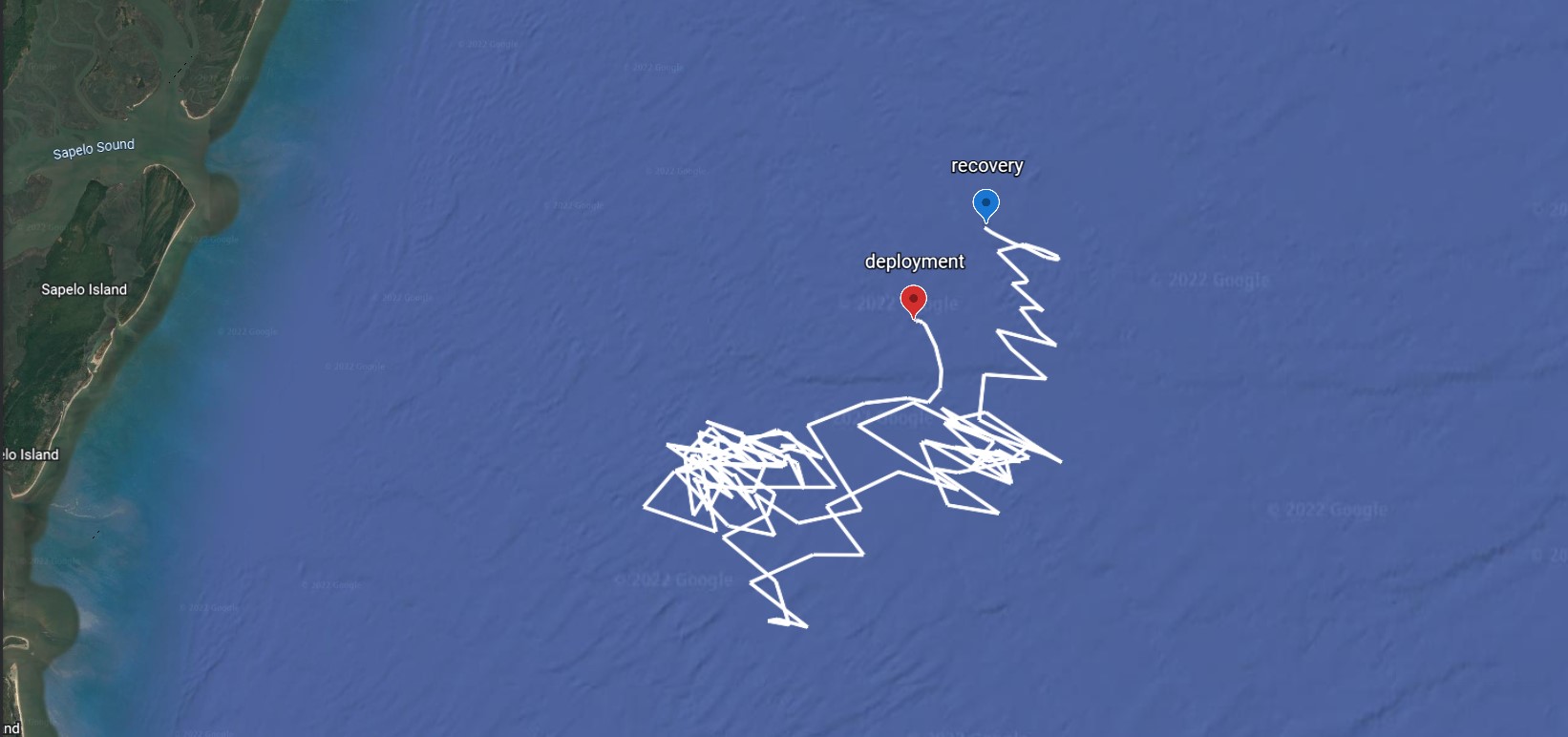}}
        \caption{Google Earth trajectory of the 2022 Angus deployment.}
        \label{angus google earth traj}
    \end{figure}

As part of an acoustic telemetry and navigation experiment, the glider Angus was deployed near Gray's Reef National Marine Sanctuary.  Its navigation was supported by the Glider Environment Networked Information system (GENIoS) \cite{chang2015real}, which generated optimal waypoints for the glider to use based on real-time glider-estimated currents and position at each surfacing.  The mission  started at the location $(31.4633^{\circ} N, 80.7757^{\circ}W)$ at April 04, 2022, 15:30 UTC, and ended at the location $(31.5072^{\circ} N, 80.7368^{\circ} W)$ at April 25, 2022, 18:51 UTC. The Google Earth trajectory of Angus is shown in Fig.~\ref{angus google earth traj}.

\section{Anomaly Detection Algorithm}
\label{anomaly detection algo}

The adopted anomaly detection algorithm can simultaneously generate glider speed estimate and flow speed estimate in real time. If the glider speed estimate is within the normal speed range, it is assumed that no anomaly has occurred. Otherwise, the glider may not be operating normally.  The algorithm-estimated flow speed is compared with glider-estimated flow speed to compute flow estimation error. If the error is too large, the detected anomaly is false alarm. This section illustrates the main framework of anomaly detection algorithm. More theoretical details can be found in the work \cite{cho2021learning}.

The glider motion model is shown in \eqref{eq 1},
\begin{gather}
    \dot{x} = F_R(x, t)  + V_R(t) \Psi_c(t) \nonumber \\
    \Psi_c(t)  = [\cos \psi_c(t), \sin\psi_c(t)]^T
\label{eq 1}
\end{gather}
where $F_R$ is the actual flow field, $x$ is the true glider position, $V_R$ is the true glider speed, and $\psi_c$ is the glider heading angle. 

In practice, the glider speed is limited due to the glider control input constraint. Equation \eqref{eq 1}  can be modified as the combination of \eqref{eq 2} and \eqref{eq 3},

\begin{gather}
    \dot{x} = F_R(x, t)  + u \Psi_c(t), 
\label{eq 2}
\end{gather}
where 
\begin{gather}
    u  =  \begin{cases} V_R, & \mbox{if } V_R \leq u_0\\
          u_0,  & \mbox{if } V_R > u_0 \end{cases}
\label{eq 3}
\end{gather}
The maximum glider speed $u_0$ is determined by the hardware configuration of the glider, as well as the water depth of operation. 
As shown in  \eqref{eq 4}, the  flow field can be represented by spatial and temporal basis functions (specifically Gaussian radial and tidal functions) \cite{liang2012real},  
\begin{gather}
    F_R(x, t) = \theta \phi(x,t)
    \label{eq 4}
\end{gather}
where \begin{gather}
     \theta = \begin{bmatrix} \theta_1 \\ \theta_2 \end{bmatrix} =  \begin{bmatrix} \theta_1^1 & \cdots & \theta_1^N \\ \theta_2^1 & \cdots & \theta_2^N \end{bmatrix} \in \mathbb{R}^{2 \times N} 
\end{gather}
is the unknown parameter to be estimated,
\begin{gather}
    \phi = \begin{bmatrix} \phi^1(x,t) & \cdots & \phi^N(x,t) \end{bmatrix}^T\\
    \phi^i(x, t) = \exp^{-\frac{||x - c_i||}{ 2 \sigma_i}} \cos(\omega_i t + \upsilon_i)
\end{gather} is the combined basis function, $c_i$ is the center, $\sigma_i$ is the width, $\omega_i$ is the tidal frequency, $\upsilon_i$ is the tidal phase, and $N$ is the selected number of basis functions.
 The heading $\Psi_c(t)$ and the true trajectory $x(t)$ are known for all time $t$ from the sbd file data operationally sent by the glider. Given the true trajectory $x(t)$, the adopted algorithm can estimate the unknown parameter $\theta$ in  \eqref{eq 4} and the glider speed $V_R$ in \eqref{eq 1} in real time. Define 
\begin{gather}
    \xi(t) = \begin{bmatrix} \xi_1(t) \\ \xi_2(t) \end{bmatrix} =  \begin{bmatrix} \xi_1^1(t) & \cdots & \xi_1^N(t) \\ \xi_2^1(t) & \cdots & \xi_2^N(t) \end{bmatrix} \in \mathbb{R}^{2 \times N} 
\end{gather}
as the estimate of the flow parameter $\theta$. Define $V_L(t)$ as the estimate of the glider speed $V_R$. The adopted algorithm can generate converged estimates, i.e.,  $\lim_{t\to \infty} \xi(t) \to \theta $ and  $\lim_{t\to \infty} V_L(t) \to V_R $ to ensure that the maximum trajectory estimation error (CLLE)  converges to zero. The adopted algorithm also designs three gains $K$, $\Bar{\gamma}_1$ and $\Bar{\gamma}_2$  to accelerate the estimating process.

 Next, the glider speed estimate $V_L(t)$ can be used to decide whether anomalies happened or not, and the flow speed estimate $\xi(t)$ can be used to avoid false alarm. The adopted algorithm assumes that the maximum glider speed $V_{max}$ and the minimum glider speed $V_{min}$ are known beforehand. If the estimated glider speed $V_L(t) \in [V_{min}, V_{max}]$, it is assumed that no anomaly happens to the glider. Otherwise, the glider may not be operating normally. In order to avoid false alarm, the adopted algorithm checks the estimated flow speed as well.  Let us introduce $F_L(t)$ as the flow speed estimated by the adopted algorithm, and $F_M(t)$ as the flow speed estimated by the glider. In practice, prior information like ocean models or sensor measurements can help generate the glider-estimated flow speed $F_M(t)$. The algorithm-estimated flow speed $F_L(t)$ is compared with the glider-estimated flow speed $F_M(t)$ to determine if the detected anomaly is false alarm or not. Defining $||F_M(t) - F_L(t)||$ as the flow estimation error, the criteria $p_E$ to evaluate the flow estimation error is given as  \eqref{eq 14},
\begin{gather}
    p_E = \frac{|| F_M(t) - F_L(t)||}{2 max (\hat{F}_{Lmax}, \hat{F}_{Mmax})}
\label{eq 14}
\end{gather}
where $\hat{F}_{Lmax} = max(||F_L(\tau)||_{\tau \in [0,t]})$ is the maximum algorithm-estimated flow speed until time $t$, and $\hat{F}_{Mmax} = max(||F_M(\tau)||_{\tau \in [0,t]})$ is the maximum glider-estimated flow speed until time $t$. If $p_E > \gamma_f$, the detected anomaly should not be trusted, where $\gamma_f$ is a false alarm threshold selected beforehand. The  pseudo-code is shown in Algorithm~\ref{algo 1}. For convenience, we introduce $f_{a}$ as anomaly detection flag shown in  \eqref{eq 15}.
\begin{gather}
     f_{a}   =  \begin{cases}  0 & \mbox{no anomaly} \\
                               1 & \mbox{anomaly detected}\\
                               2 & \mbox{false alarm}
         \end{cases}
\label{eq 15}
\end{gather}

\begin{algorithm}
\caption{Anomaly Detection}
\label{algo 1}
 \hspace*{\algorithmicindent} \textbf{Input:} Algorithm-estimated flow speed $F_L(t)$, glider-estimated flow speed $F_M(t)$, false alarm threshold $\gamma_f$, estimated glider speed $V_L(t)$, maximum glider speed $V_{max}$, minimum glider speed $V_{min}$ \hspace*{\algorithmicindent}  \textbf{Output:}  Anomaly detection flag $f_a$
\begin{algorithmic}[1]
\STATE $\hat{F}_{Lmax} = max(||F_L(\tau)||_{\tau \in [0,t]})$
\STATE $\hat{F}_{Mmax} = max(||F_M(\tau)||_{\tau \in [0,t]})$
\STATE  $  p_E = \frac{|| F_M(t) - F_L(t)||}{2 max (\hat{F}_{Lmax}, \hat{F}_{Mmax})} $
\IF {$p_E > \gamma_f $} 
  \STATE $f_a = 2$
\ELSIF{$V_L(t) > V_{max}$ or $V_L(t) < V_{min}$}
   \STATE $f_a = 1$
 \ELSE
    \STATE $f_a = 0$
\ENDIF 
\end{algorithmic}
\end{algorithm}

\section{Experimental Results}
\label{experimental results}
For the purpose of validation, the adopted anomaly detection algorithm is applied to two field deployments with verified anomalies in the coastal ocean of the Gulf of Mexico and south east Atlantic Ocean. The estimated glider speed is compared with the maximum and minimum glider speeds to decide if anomalies can be identified. The estimated flow speed is checked against the glider-estimated flow speed to avoid false alarm. For verification, the anomaly detected by the algorithm is compared with the anomaly seen from glider dbd file data and pilot logs. For reference, the designed parameters of the algorithm are shown in TABLE~\ref{table parameters}. 
\begin{table}[htbp]
\caption{Parameters of Experiments}
\begin{center}
\begin{tabular}{| m{2cm} || m{2cm} | m{2cm} |}
 \hline
 Parameters & Stella Deployment Data & Angus Deployment Data\\
 \hline
  number of basis functions N  & 4 & 4 \\
 \hline
  width $\sigma_i$   & 1.3e4    & 2e4 \\
 \hline
  tidal phase $\upsilon_i$ &   0  & 0  \\
   \hline
  tidal frequency $\omega_i$ & $2\pi$e-6 &  $2\pi$e-6 \\
 \hline 
  gain K &  $\begin{bmatrix} 0.003 & 0 \\  0 & 0.003 \end{bmatrix}$ & $\begin{bmatrix} 0.002 & 0 \\  0 & 0.002 \end{bmatrix}$ \\ 
  \hline
  gain $\Bar{\gamma}_1$  & 1e-7 & 2e-5 \\
  \hline
  gain $\Bar{\gamma}_2$ & 35e-10 & 2e-8  \\
  \hline
  false alarm threshold $\gamma_f$ & 0.6 & 0.7\\
  \hline
\end{tabular}
\label{table parameters}
\end{center}
\end{table}

\subsection{Results of Stella Deployment}

\begin{figure}[htbp]
     \centering
     \begin{subfigure}[b]{0.3\textwidth}
         \centerline{\includegraphics[width=\textwidth]{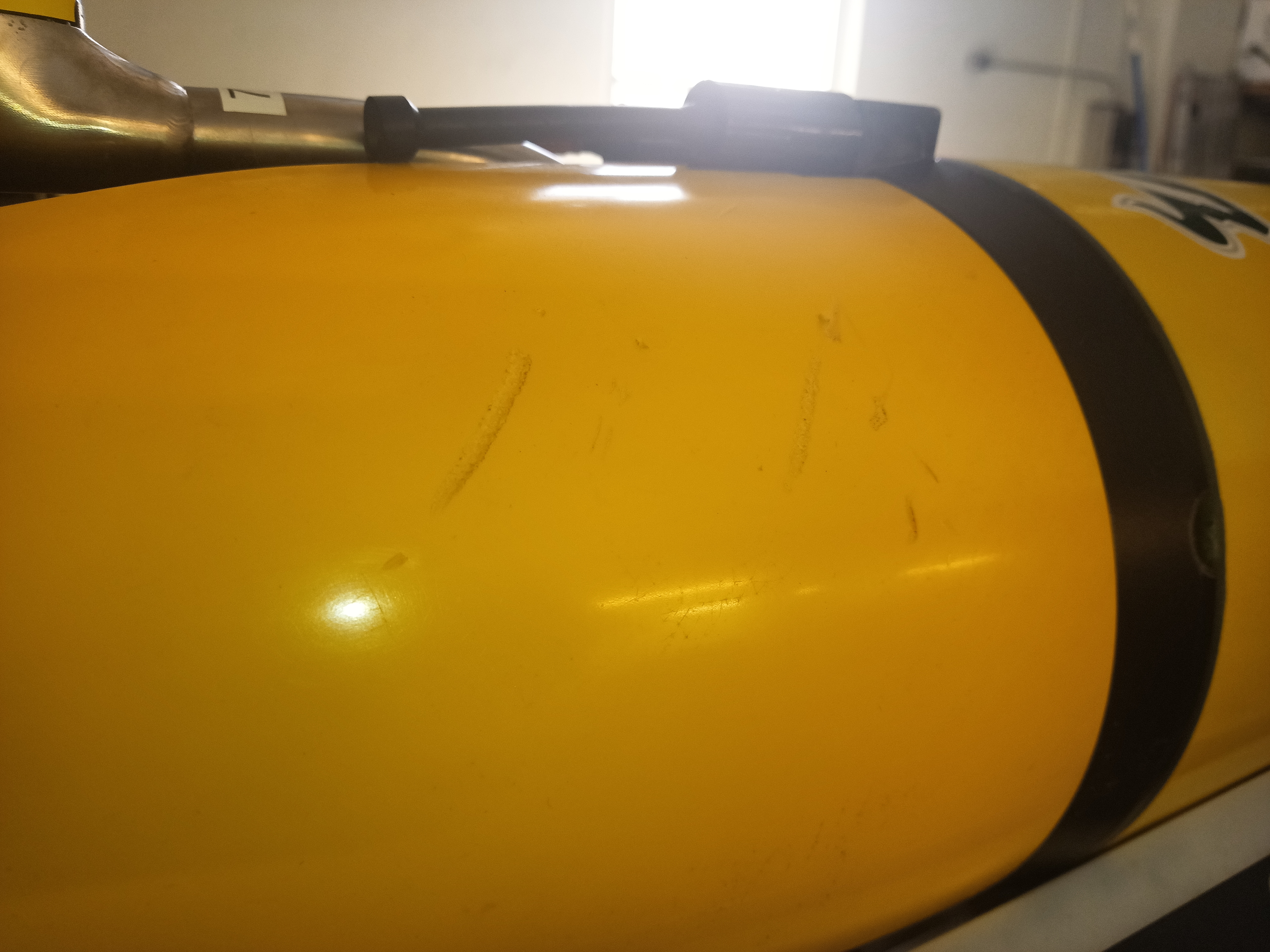}}
         \caption{Hull scratches documented after recovery.}
         \label{shark scratch}
     \end{subfigure}
     \hfill
     \begin{subfigure}[b]{0.3\textwidth}
         \centerline{\includegraphics[width=\textwidth]{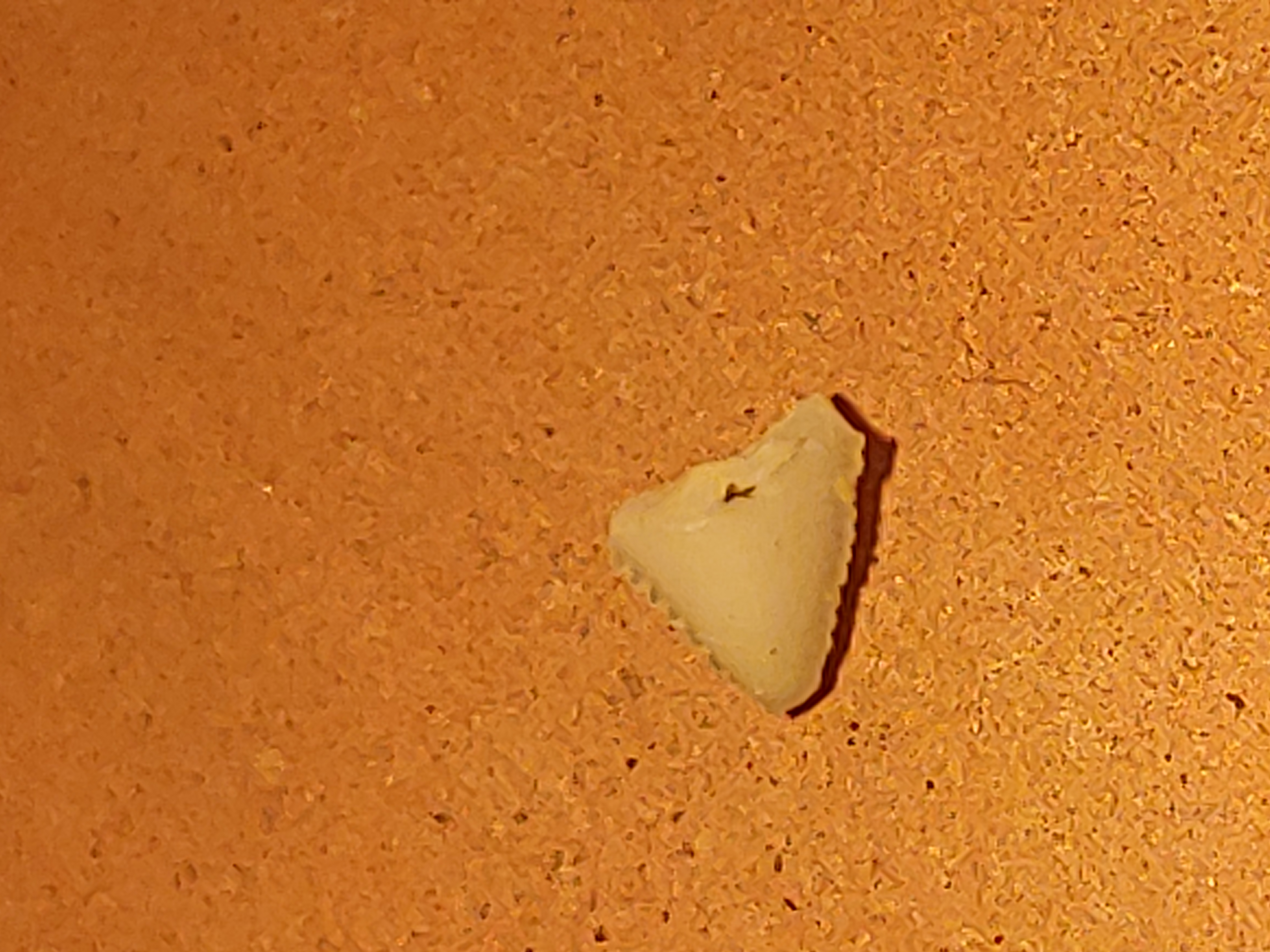}}
         \caption{Shark tooth found embedded in the hull after recovery.}
         \label{shark tooth}
     \end{subfigure}
     
        \caption{Post-recovery evidence of shark strike. }
        \label{shark hit evidence}
    \end{figure}

\begin{figure}[htbp]
     \centering
     \begin{subfigure}[b]{0.4\textwidth}
         \centerline{\includegraphics[width=\textwidth]{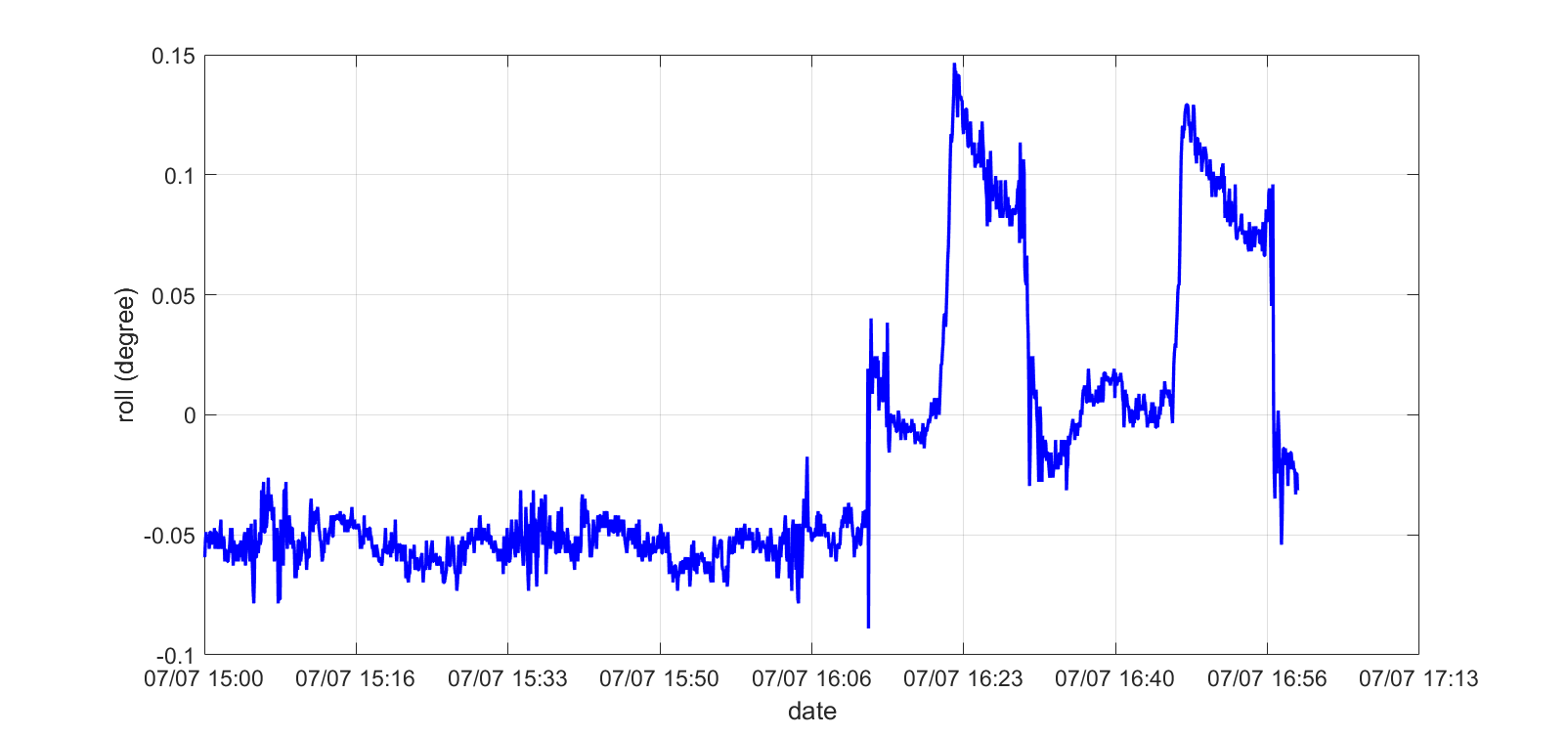}}
         \caption{Glider-measured roll (degrees) from post-recovery DBD data.}
         \label{shark hit roll}
     \end{subfigure}
     \hfill
     \begin{subfigure}[b]{0.4\textwidth}
         \centerline{\includegraphics[width=\textwidth]{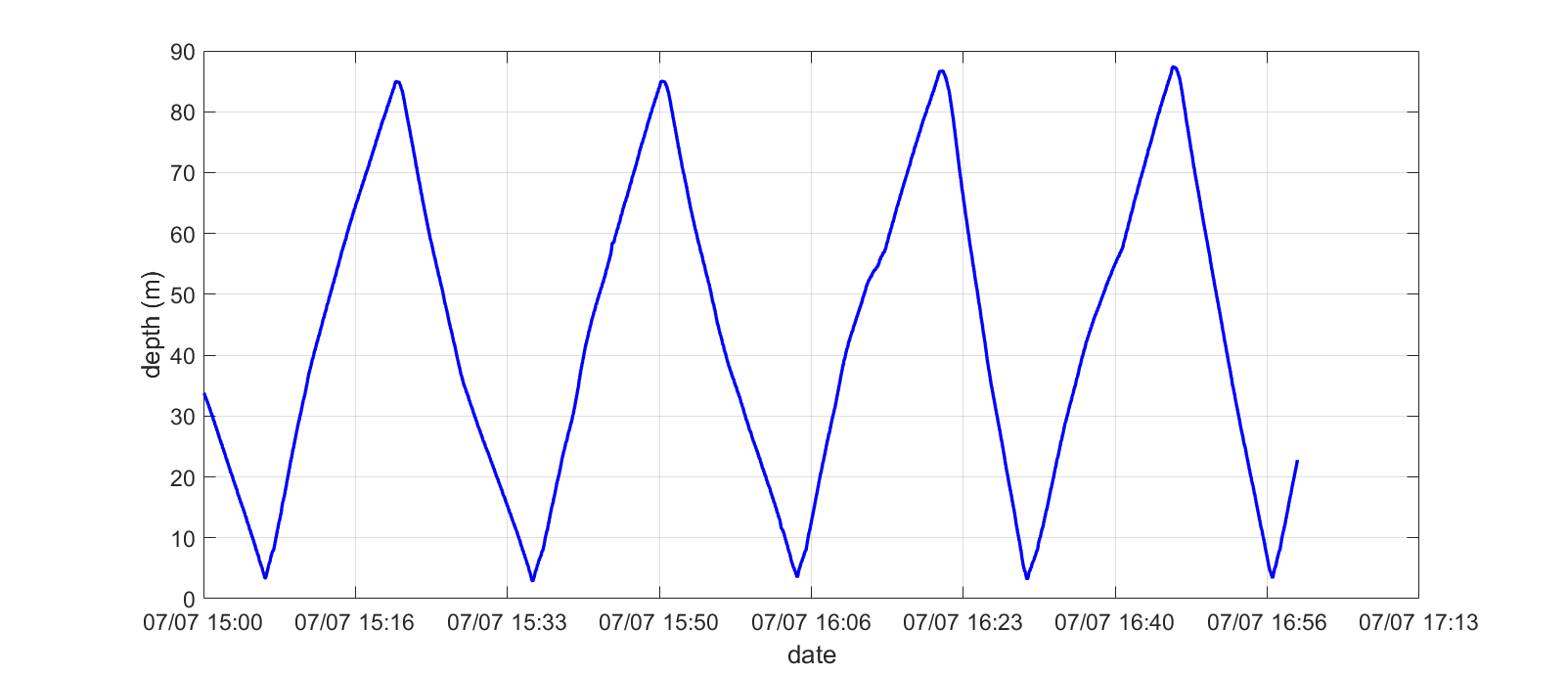}}
         \caption{Glider-measured depth (m) from post-recovery DBD data.}
         \label{shark hit depth}
     \end{subfigure}
        \caption{ground truth (Stella deployment).}
        \label{shark hit ground truth data}
    \end{figure}

Upon recovery July 19th, 2021, the glider Stella missing one of its two detachable wings, and scratches on the glider hull and the shark tooth embedded in the thermoplastic aft glider cowling (Figs.~ref{shark scratch} and \ref{shark tooth}) suggest a serious shark strike.  Glider pilot analysis of the full DBD data post deployment suggests that the shark hit of the glider likely occurred around July 07, 2021, 16:00 UTC when it was descending at about 55m depth. This post-mission analysis shows that the glider went from a $-4 deg$  roll to an $8 deg$ roll, and finally settled back at about a $0 deg$ roll (Fig.~\ref{shark hit roll} and Fig.~\ref{shark hit depth}).  The sudden change of roll and the consistent change through the remainder of the deployment suggests that the wing was lost during this process.  

\begin{figure}[htbp]
        \centerline{\includegraphics[width=0.4\textwidth]{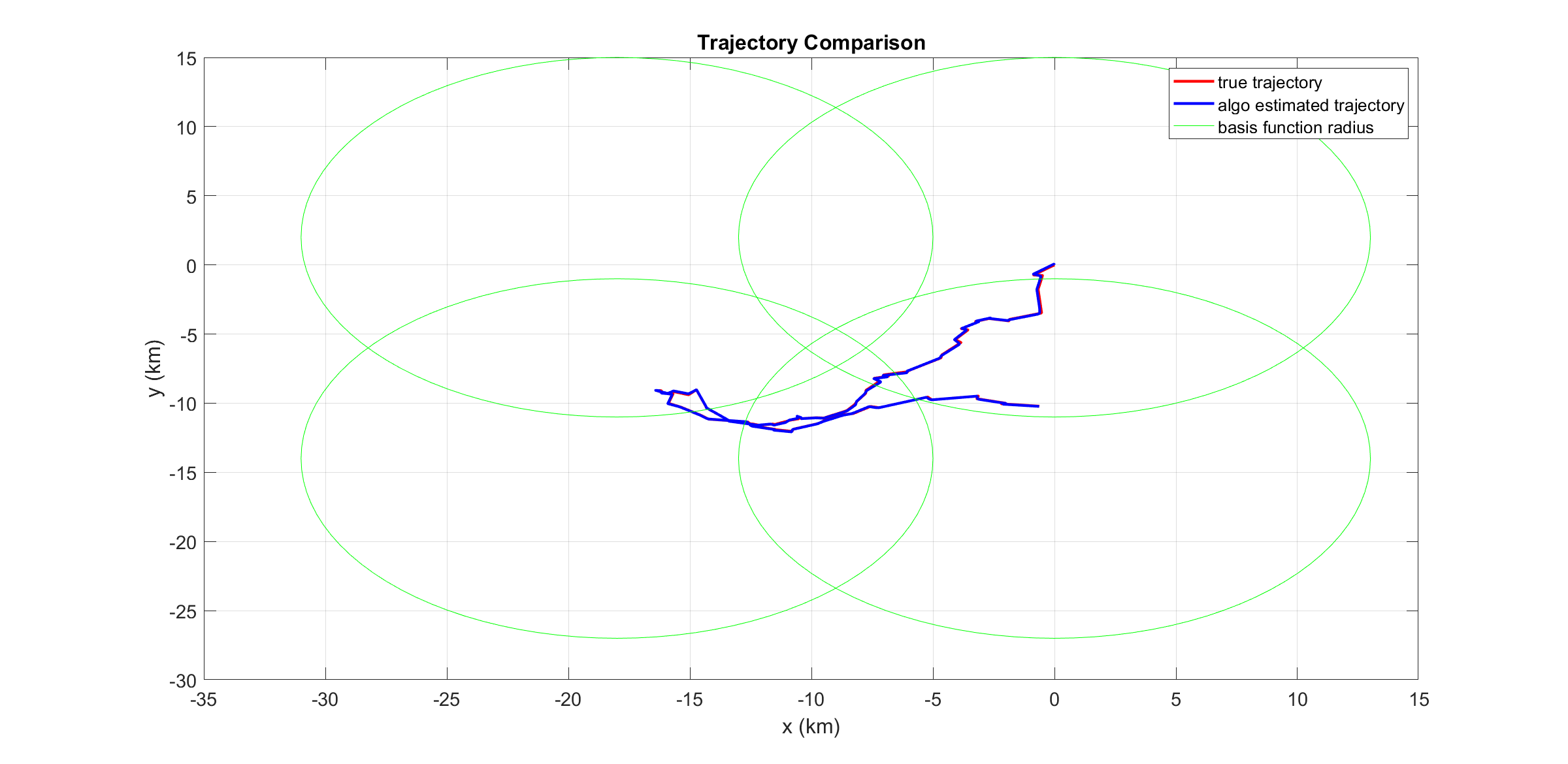}}
          \caption{Comparison of the estimated (blue) and reported ``true" (red) trajectory  for the 2021 Stella deployment.}
        \label{shark hit trajectory comparison}
    \end{figure}

\begin{figure}[htbp]
        \centerline{\includegraphics[width=0.4\textwidth, height = 4cm]{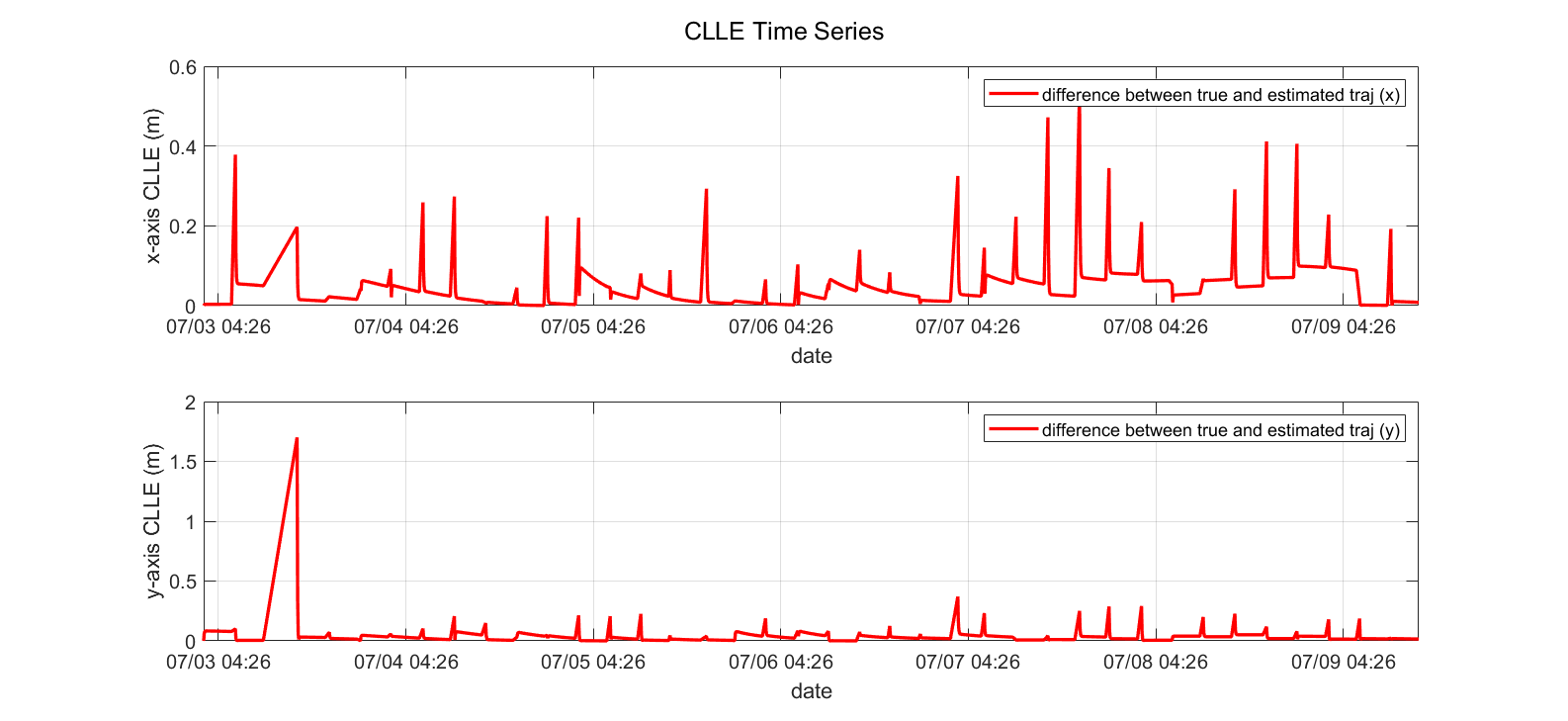}}
        \caption{CLLE (m) for the 2021 Stella deployment.}
        \label{shark hit CLLE}
    \end{figure}

Applying the anomaly detection algorithm to the telemetered SBD data from the glider, the adopted algorithm can guarantee that the estimated trajectory converges to the true trajectory since the persistent excitation matrix $W(t)$ is not singular. The comparison of the estimated trajectory with the true trajectory is shown in Fig.~\ref{shark hit trajectory comparison}. It should be noted that the 4 green circles cover the whole trajectory, which means that the 4 combined basis functions cover the whole flow fields. This coverage is necessary for the parameter estimation convergence. As shown in Fig.~\ref{shark hit CLLE}, the maximum trajectory estimation error (CLLE) is only  $1.7m$, which is small enough considering the glider moved approximately 0.25m/s over O(100)km in its deployment. Therefore, it can be concluded that the CLLE converges to zero.
    
\begin{figure}[htbp]
     \centering
     \begin{subfigure}[b]{0.4\textwidth}
         \centerline{\includegraphics[width=\textwidth, height = 3cm]{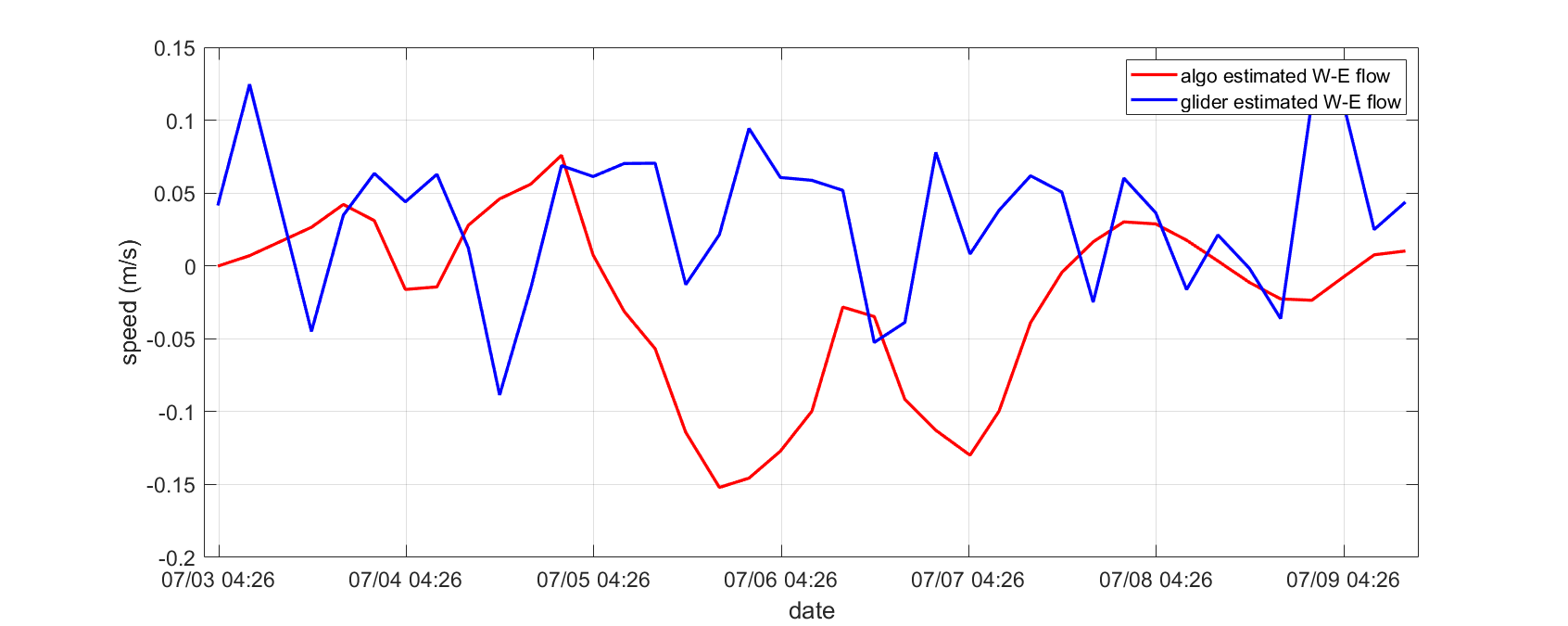}}
           \caption{W-E flow component.}
         \label{shark hit W-E flow}
     \end{subfigure}
     \hfill
     \begin{subfigure}[b]{0.4\textwidth}
         \centerline{\includegraphics[width=\textwidth, height = 3cm]{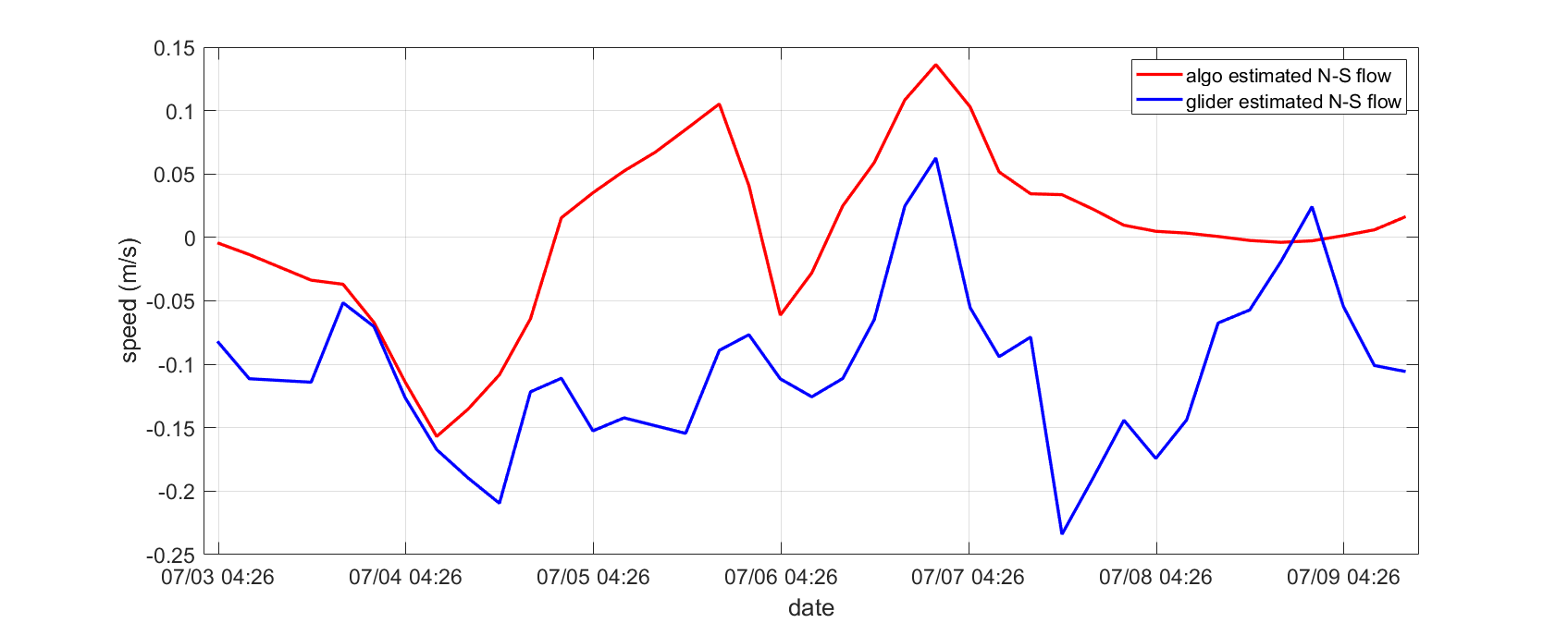}}  
           \caption{N-S flow component.}
         \label{shark hit N-S flow}
     \end{subfigure}
     
      \caption{Comparison of glider-estimated and algorithm-estimated W-E ($u$, upper) and N-S ($v$, lower) velocities. }
        \label{shark hit flow comparison}
    \end{figure}

Given the CLLE convergence,  the glider speed estimate and the flow speed estimate should also converge. In this case, the anomaly can be detected based on these two estimates. For better comparison, the flow is divided into $u$ (West-East, W-E, or zonal) and $v$ (North-South, N-S, or meridional) components. As shown in Fig.~\ref{shark hit W-E flow}, the algorithm-estimated W-E flow is close to the corresponding glider-estimated W-E flow, which means that the $u$ flow estimation error is small. Similarly as the N-S flow shown in Fig.~\ref{shark hit N-S flow}.  This comparison suggests that the anomaly detection flag can be trusted when it is signaled.
    
\begin{figure}[htbp]
        \centerline{\includegraphics[width=0.4\textwidth]{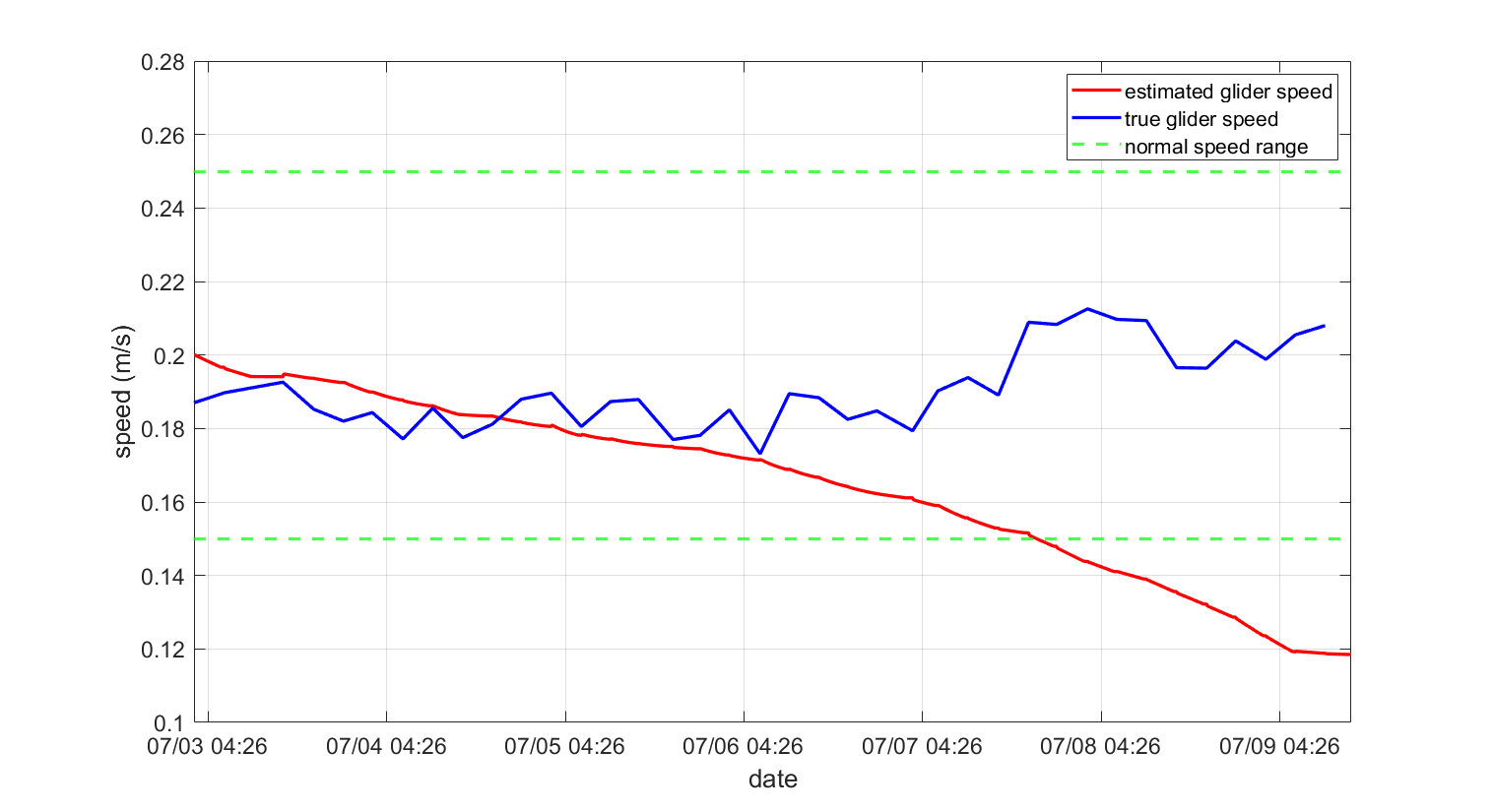}}
        \caption{Comparison of estimated (red) and reported ``true" (blue) speed of Stella in the 2021 deployment.}
        \label{shark hit speed comparison}
    \end{figure}

\begin{figure}[htbp]
        \centerline{\includegraphics[width=0.4\textwidth, height = 3.5cm]{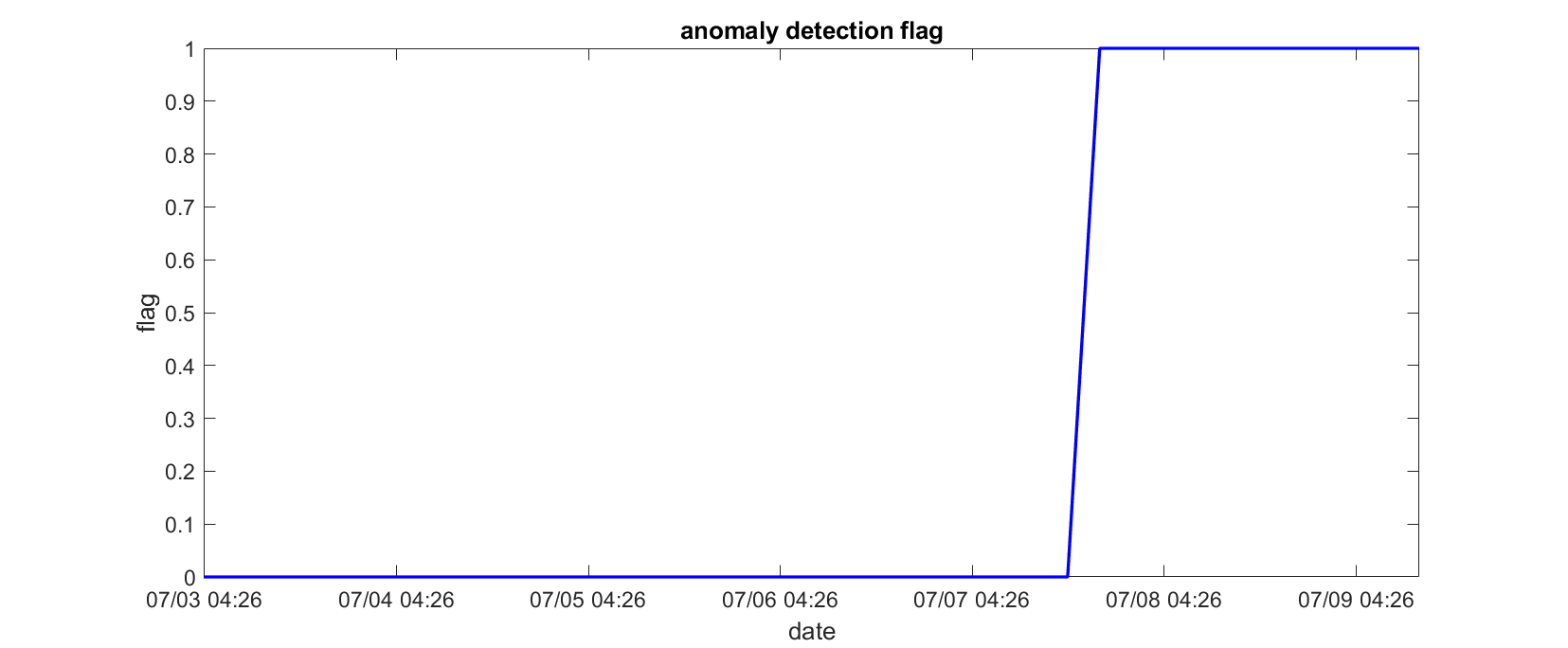}}
        \caption{Anomaly detection flag as calculated by the adopted algorithm for the 2021 Stella deployment.}
        \label{shark hit flag}
    \end{figure}

The estimated glider speed is compared with the reported ``true" speed value collected by the glider. If the estimated speed is not in the normal speed range, the anomaly is assumed to have occurred. As shown in Fig.~\ref{shark hit speed comparison},  the estimated glider speed drops out of the normal speed range (green dot line) at around July 07, 2021, 18:00 UTC, and keeps dropping afterwards. As shown in Fig.~\ref{shark hit flag}, the flag value switches from 0 to 1 at around July 07, 2021, 18:00 UTC as well. The flag value never changes to 2 with the false alarm threshold $\gamma_f = 0.6$. The timestamp around which the anomaly is detected by the adopted algorithm corresponds to the timestamp from both the glider team's report and the full post-deployment DBD file data. Therefore, the adopted algorithm is verified valid and accurate by successfully detecting the anomaly.

\subsection{Results of Angus Deployment}
\begin{figure}[htbp]
        \centerline{\includegraphics[width=0.4\textwidth, height=4cm]{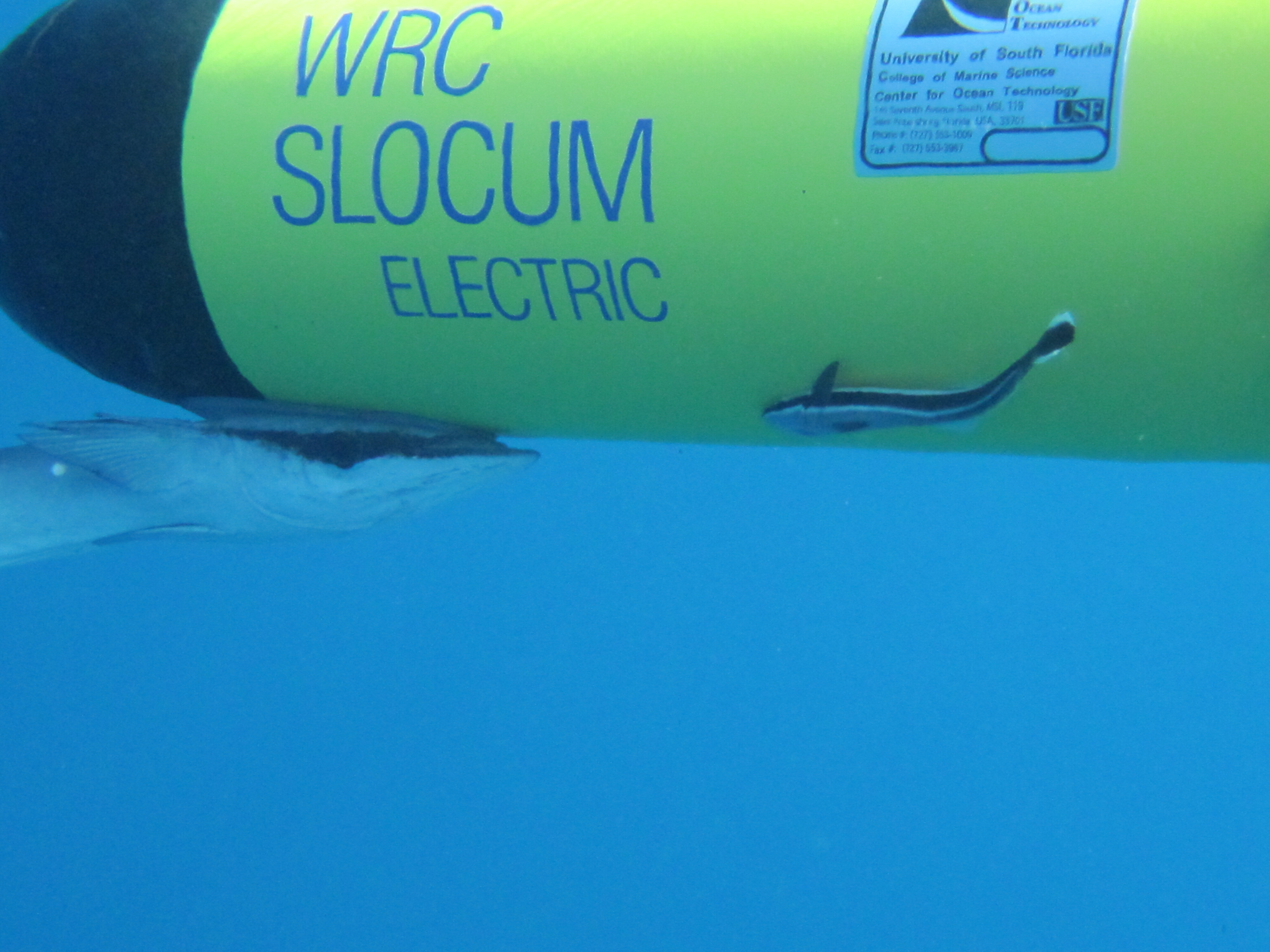}}
        \caption{Example image of remora attachment (photo credit: Chad Lembke, College of Marine Science, University of South Florida).}
        \label{remora}
    \end{figure}

\begin{figure}[htbp]
     \centering
     \begin{subfigure}[b]{0.4\textwidth}
         \centerline{\includegraphics[width=\textwidth]{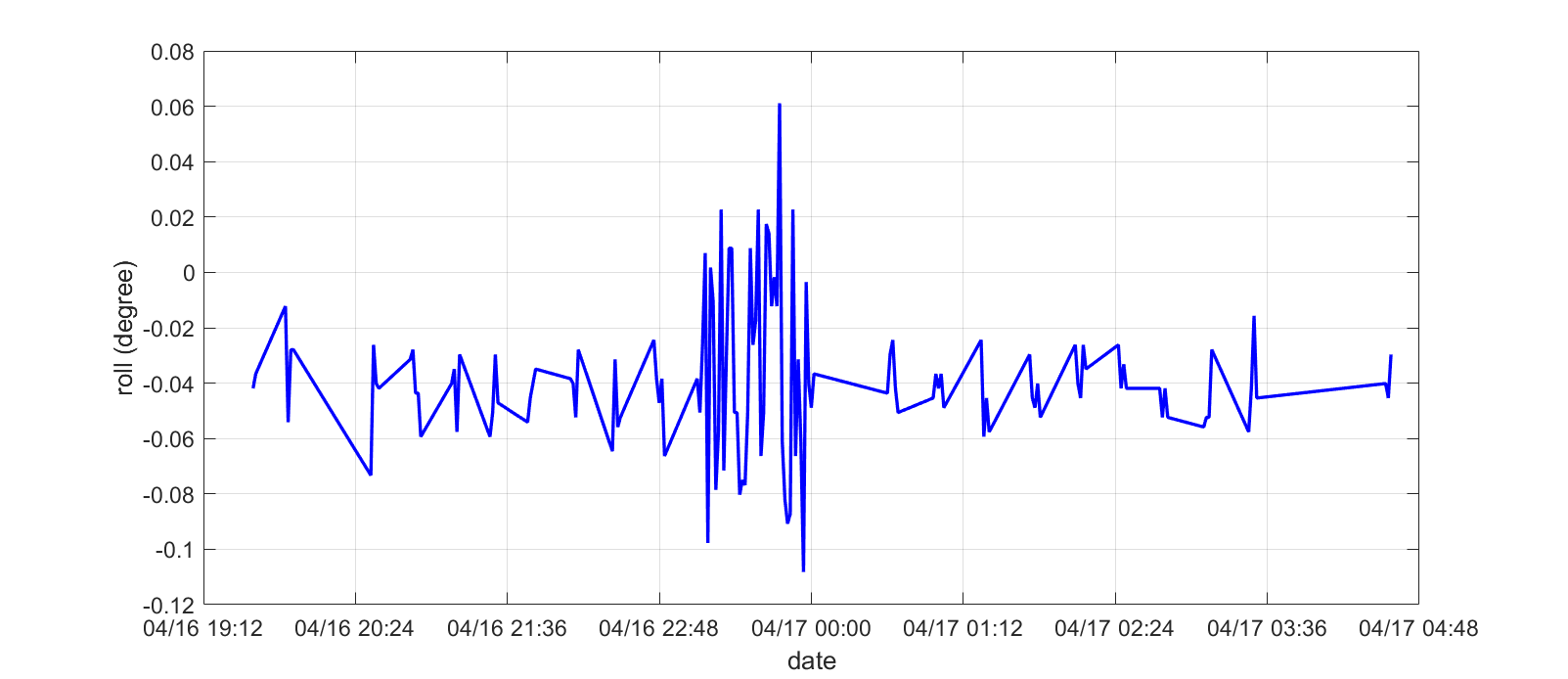}}
         \caption{Glider-measured roll (degrees) from post-recovery DBD data.}
         \label{remora roll}
     \end{subfigure}
     \hfill
     \begin{subfigure}[b]{0.4\textwidth}
         \centerline{\includegraphics[width=\textwidth]{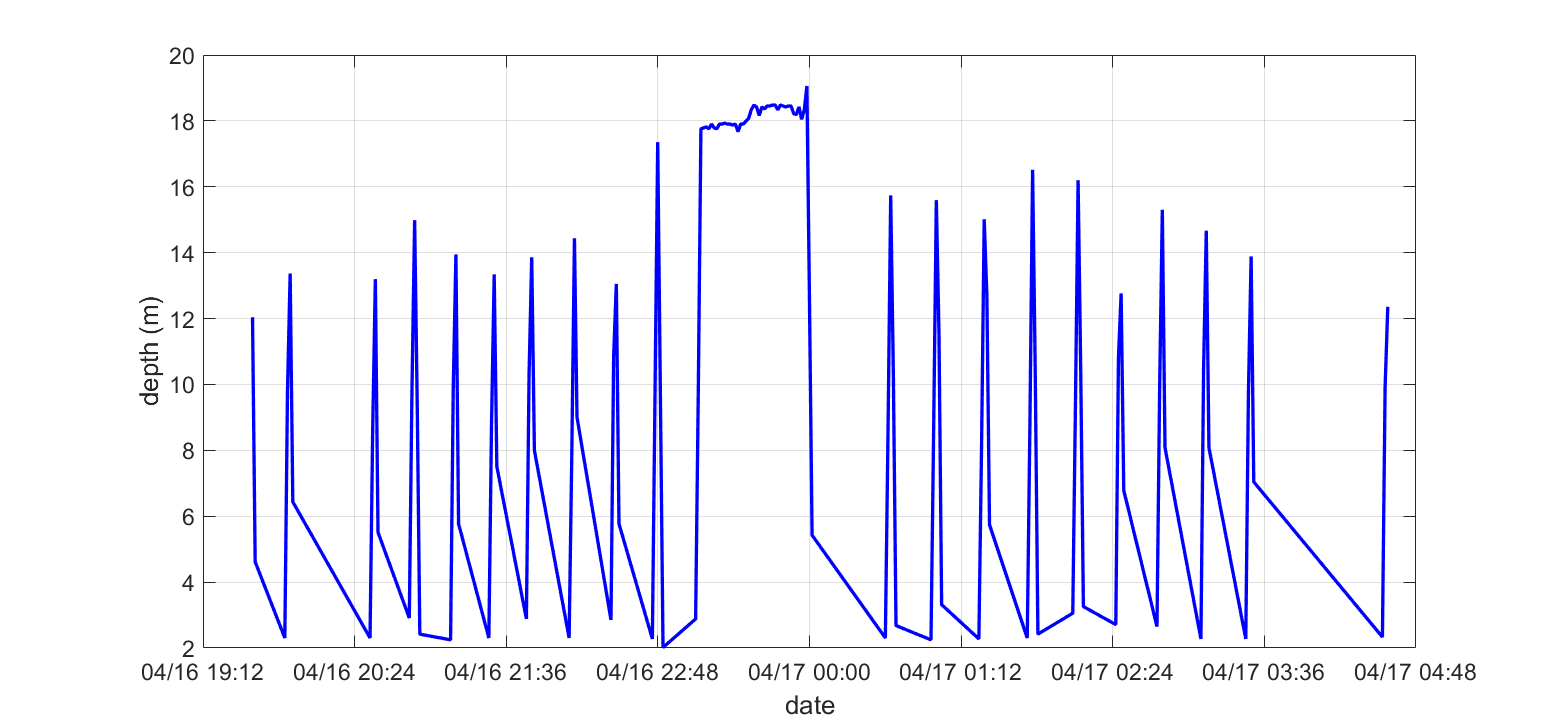}}
         \caption{Glider-measured depth (m) from post-recovery DBD data.}
         \label{remora depth}
     \end{subfigure}

        \caption{ground truth (Angus deployment).}
        \label{shark hit ground truth data}
   \end{figure}

According to the mission report, pilots noted that the glider Angus periodically flew slower than expected, and suffered from unusually poor satellite data transmission, with multiple broken connections per surfacing, consistent with remora strikes.  The problem was first identified by pilots around April 17, 2022, 00:00 UTC. Since remoras are negatively buoyant, the glider's Iridium antenna is not fully out of the water, and multiple animals attached to the instrument can induce too much of a buoyancy loss for the glider's pump to overcome, preventing flight altogether.  The remora attachment scene is difficult to catch in real-time missions. For illustration, Fig.~\ref{remora} shows an example image of two remoras attached to a glider (photo credit: Chad Lembke, College of Marine Science, University of South Florida). Based on the post-recovery DBD file data, the roll and depth ground truth is shown in Fig.~\ref{remora roll} and Fig.~\ref{remora depth}, respectively. The timestamp April 17, 2022, 00:00 UTC when depth and roll data becomes abnormal exactly matches the glider team's reported timestamp. 

\begin{figure}[htbp]
        \centerline{\includegraphics[width=0.4\textwidth, height = 4cm]{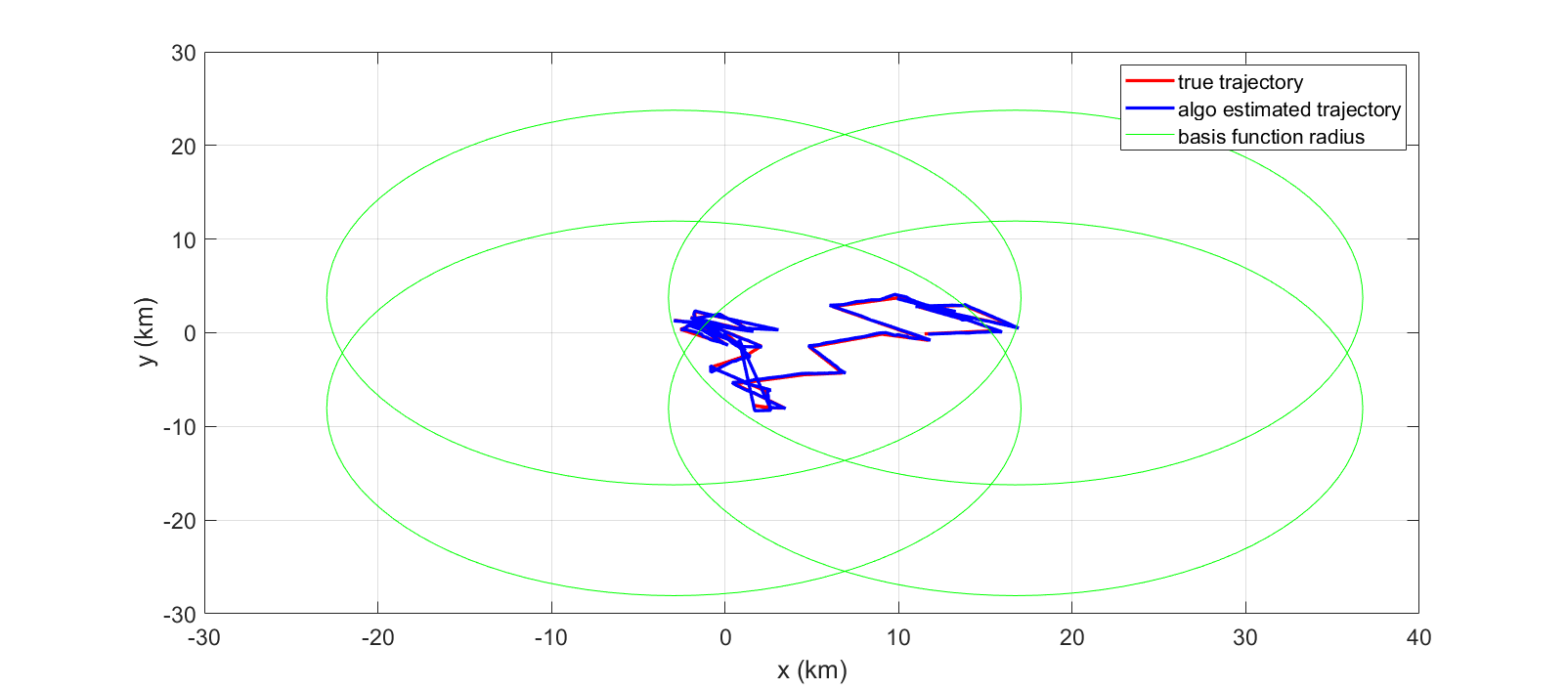}}
        \caption{Comparison of the estimated (blue) and reported ``true" (red) trajectory  for the 2022 Angus deployment.}
        \label{remora trajectory comparison}
    \end{figure}

\begin{figure}[htbp]
        \centerline{\includegraphics[width=0.4\textwidth, height = 4cm]{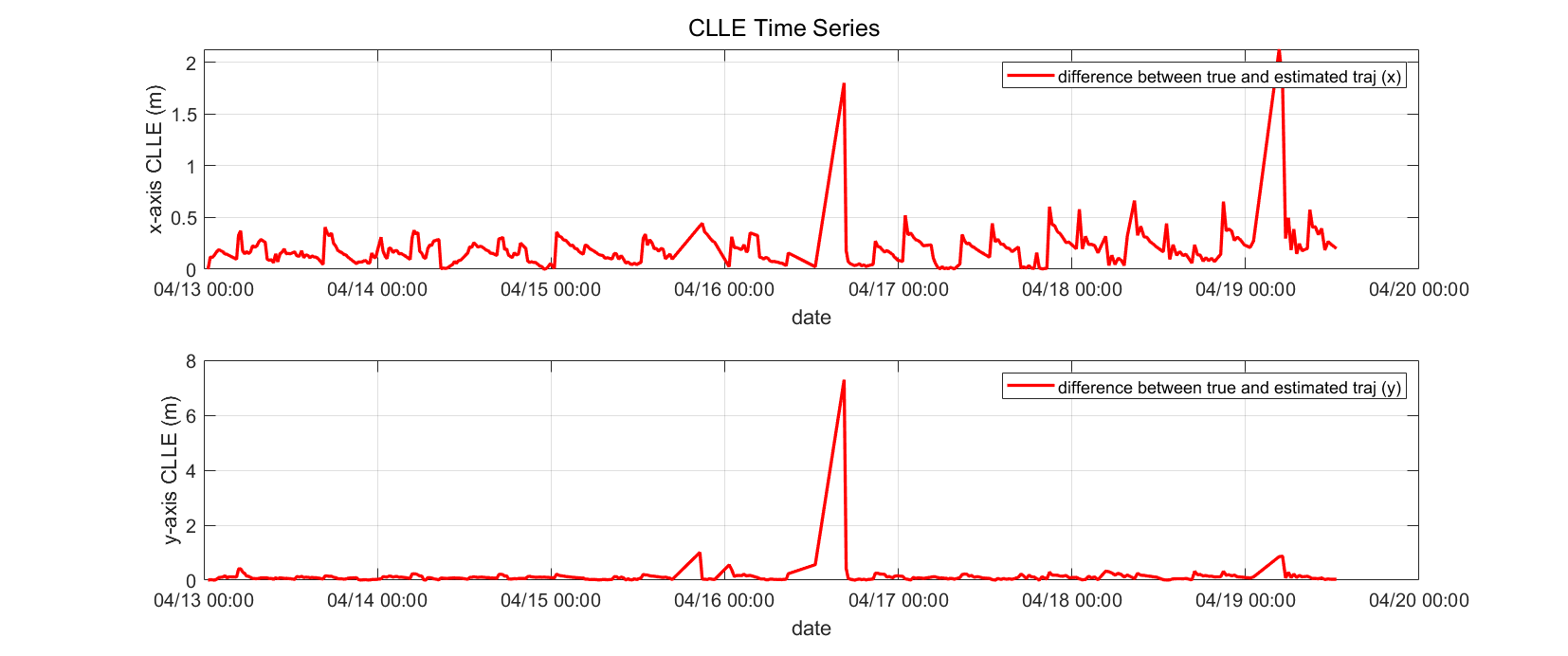}}
        \caption{CLLE (m) for the 2022 Angus deployment.}
        \label{remora CLLE}
    \end{figure}

Using the SBD file offloaded from the glider, the calculated persistent excitation matrix $W(t)$ is not singular, which means that the adopted algorithm can guarantee that the estimated trajectory converges to the true trajectory. The comparison of the estimated trajectory with the true trajectory is shown in Fig.~\ref{remora trajectory comparison}. It should be noted that the four green circles cover the whole trajectory, which means that the four combined basis functions cover the whole flow fields. This coverage is necessary for the parameter estimation convergence. As shown in Fig.~\ref{remora CLLE}, the maximum trajectory estimation error (CLLE) only reaches around $6m$, which is sufficiently small given the glider speed and range.  Therefore, it can be concluded that the CLLE converges to zero.
    
\begin{figure}[htbp]
     \centering
     \begin{subfigure}[b]{0.4\textwidth}
         \centerline{\includegraphics[width=\textwidth, height = 3cm]{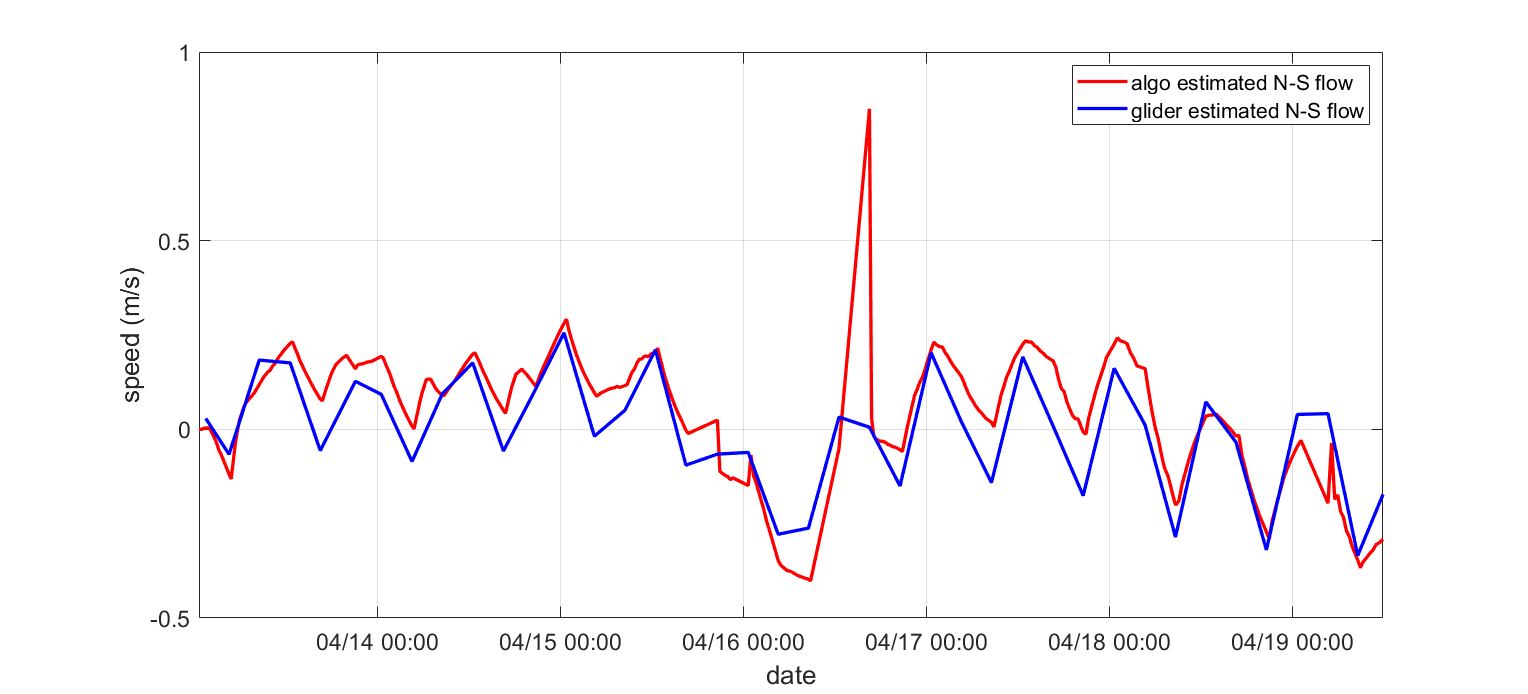}}
         \caption{W-E flow component.}
         \label{remora W-E flow}
     \end{subfigure}
     \hfill
     \begin{subfigure}[b]{0.4\textwidth}
         \centerline{\includegraphics[width=\textwidth, height = 3cm]{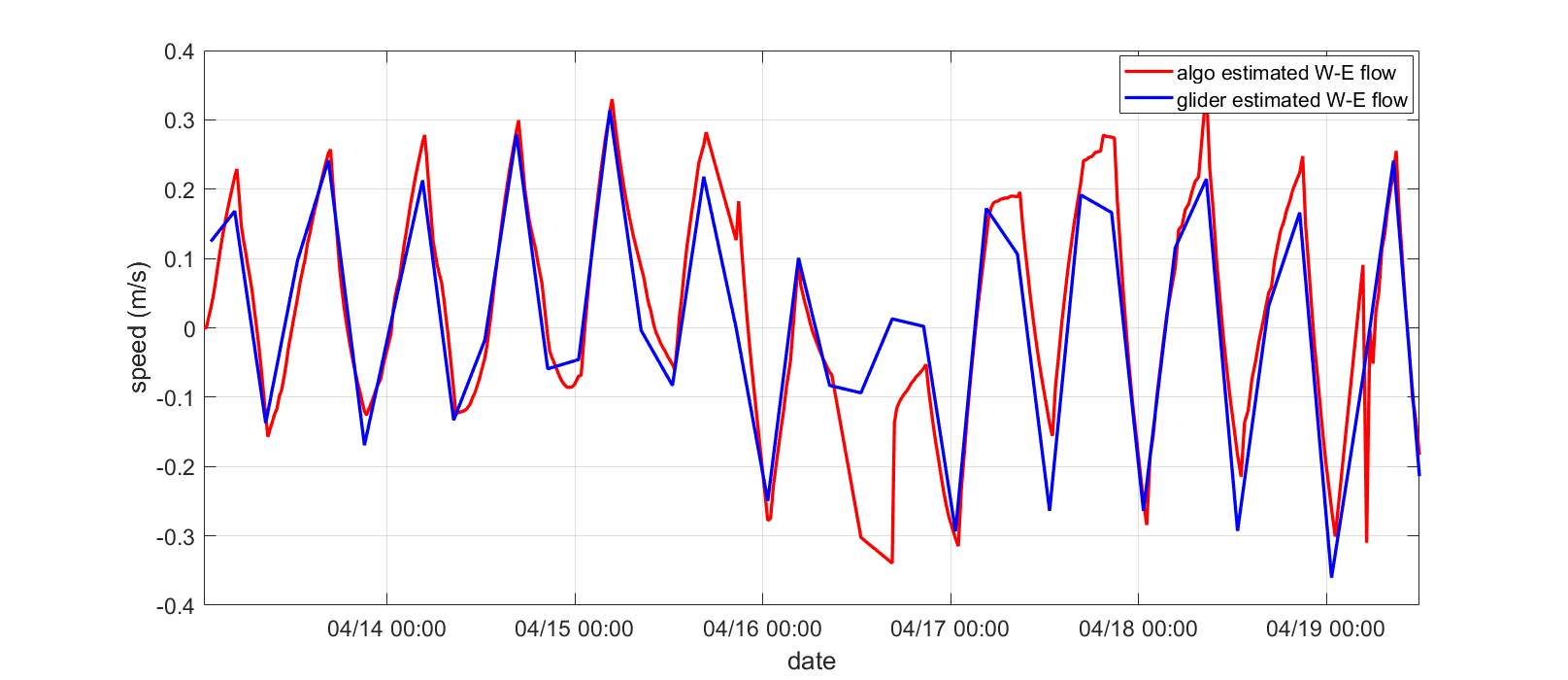}}
         \caption{N-S flow component.}
         \label{remora N-S flow}
     \end{subfigure}
 
      \caption{Comparison of glider-estimated and algorithm-estimated W-E ($u$, upper) and N-S ($v$, lower) velocities. }
        \label{remora flow comparison}
    \end{figure}

Given the CLLE convergence,  the glider speed estimate and the flow speed estimate should  converge as well. In this case, the anomaly can be detected based on these two estimates. For better comparison, the flow is divided into $u$ (West-East, W-E, or zonal) and $v$ (North-South, N-S, or meridional) components. As shown in Fig.~\ref{remora W-E flow} and ~\ref{remora N-S flow}, the algorithm-estimated W-E and N-S flow is close to the corresponding glider-estimated values, so both the $u$ and $v$ flow estimation errors are small. Therefore, any anomaly detection flag can be trusted.  

\begin{figure}[htbp]
        \centerline{\includegraphics[width=0.4\textwidth, height = 3.5cm]{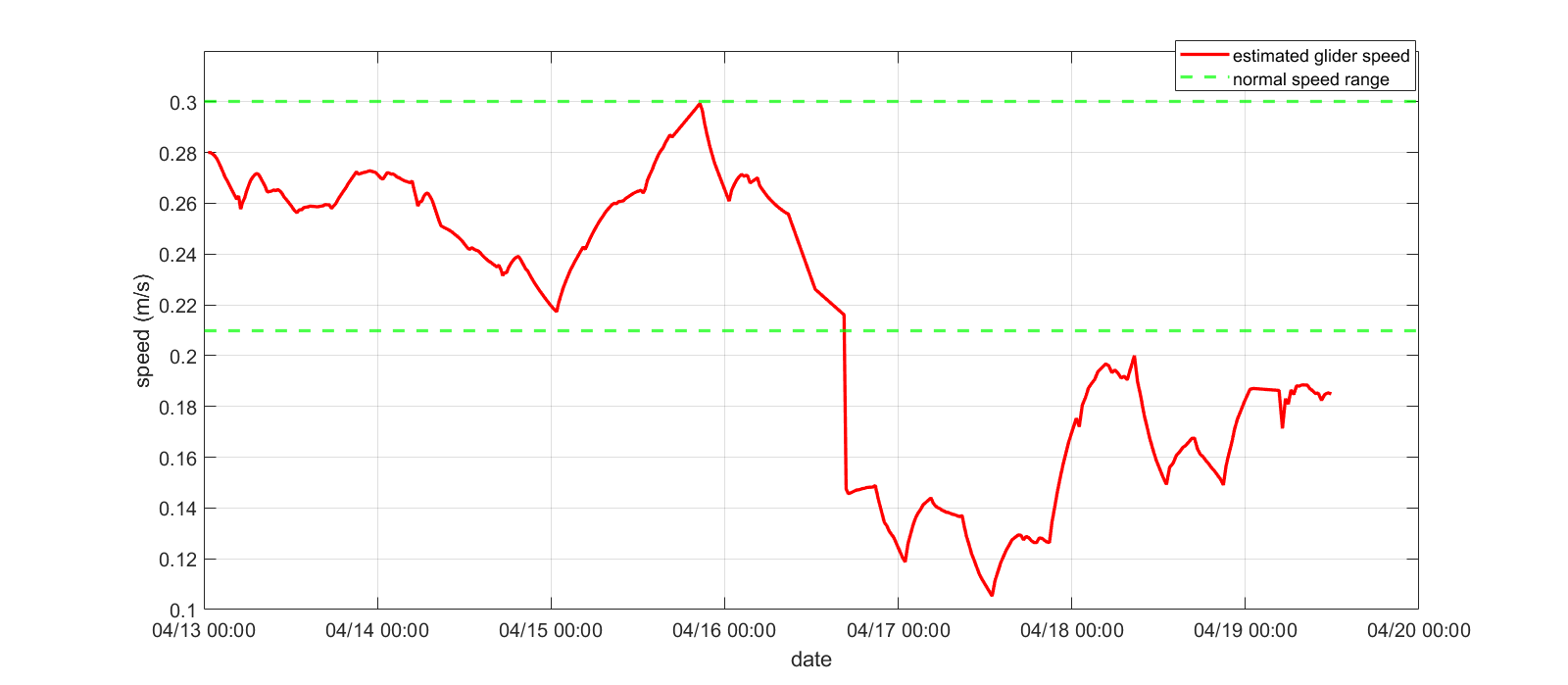}}
        \caption{Comparison of estimated speed (red) and normal speed range (green) of Angus in the 2022 deployment.}
        \label{remora glider speed}
    \end{figure}

\begin{figure}[htbp]
        \centerline{\includegraphics[width=0.4\textwidth, height = 3.5cm]{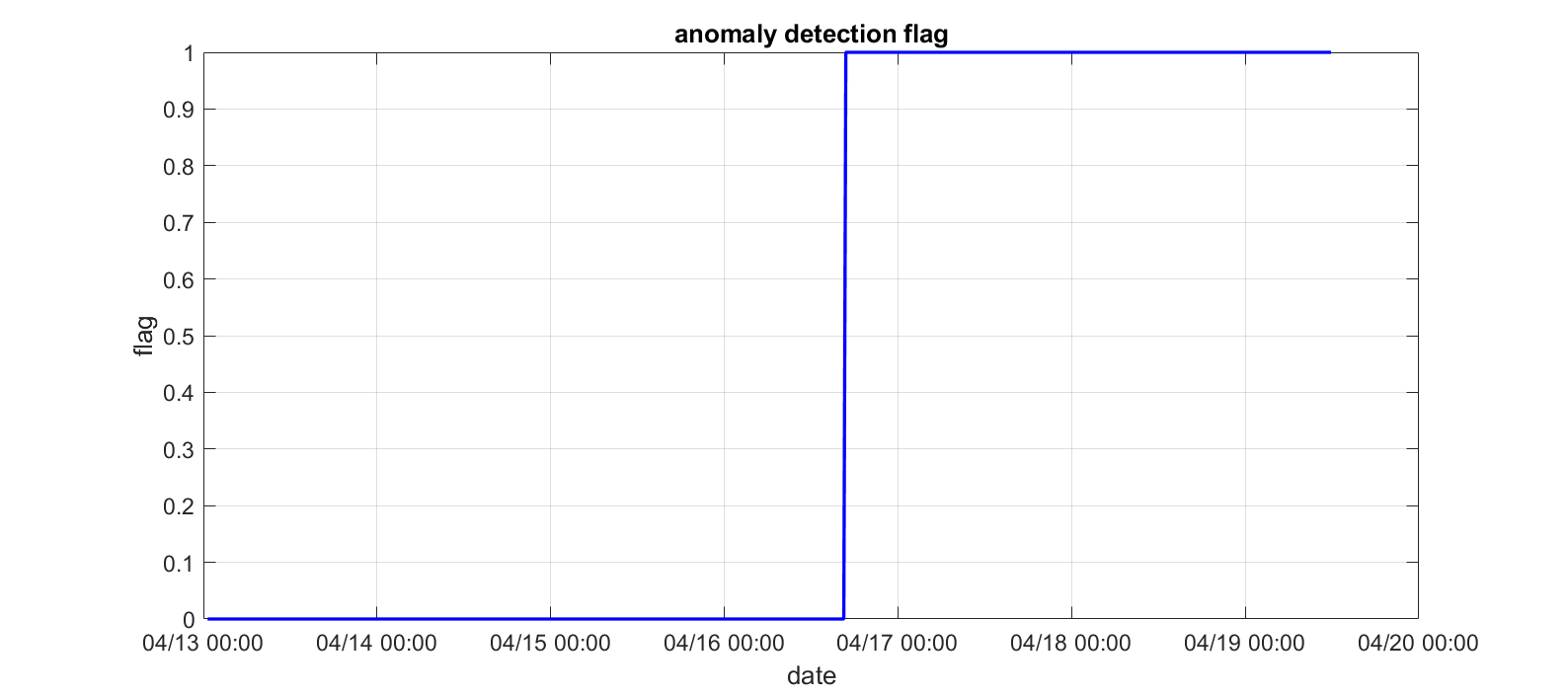}}
         \caption{Anomaly detection flag as calculated by the adopted algorithm for the 2022 Angus deployment.}
        \label{remora flag}
    \end{figure}

As shown in Fig.~\ref{remora glider speed},  the estimated glider speed sharply drops out of the normal speed range(green dot line) at around April 16, 2022, 22:00 UTC. Afterwards, the glider tries to recover its speed but continues below the normal range. As shown in Fig.~\ref{remora flag}, the flag value switches from 0 to 1 at around April 16, 2022, 22:00 UTC as well. The flag value never changes to 2 with the false alarm threshold $\gamma_f = 0.7$. The timestamp around which the anomaly is detected by the adopted algorithm corresponds to the timestamp from both the glider team's report and the DBD file data. Therefore, the adopted algorithm is verified valid and accurate, and the anomaly is successfully detected.

\section{Conclusion}
\label{conclusion}
This paper implements and validates an anomaly detection algorithm for two sets of data in real missions of gliders, the University of South Florida (USF) glider Stella and the Skidaway Institute of Oceanography (SkIO) glider Angus, where the anomalies of shark attack and remora attachment can be verified. Based on the glider trajectory and the heading angle data, the algorithm generates glider speed estimate and flow speed estimate in real time. The estimated glider speed is compared with the normal glider speed range to decide whether anomalies happen or not. The estimated flow speed is compared with the glider-estimated flow speed to determine if the detected anomaly is a false alarm. From a theoretical point of view, the current algorithm focuses on the real-time estimation process by predicting and updating the trajectory, which is known as the filtering framework, as the ongoing deployment gives the real glider data as feedback. Future work will improve the estimation accuracy by integrating the post-deployment glider data as a smoothing framework. From a practical point of view, future work will implement algorithms that account for dependence of glider speed on water depth, and consider the dependency of error on glider heading angle, which can be used to diagnose poor ballasting of the instrument.  
\bibliography{IEEEabrv, reference.bib}
\bibliographystyle{IEEEtran}

\end{document}